%% file: main.tex
\documentclass{article}

\usepackage{arxiv}
\input{math_commands.tex}

\usepackage[utf8]{inputenc}
\usepackage[T1]{fontenc}
\usepackage{amsmath}
\usepackage{amssymb}
\usepackage{array}
\usepackage{ragged2e}
\usepackage{afterpage}
\usepackage{placeins}
\usepackage{float}
\usepackage{booktabs}
\usepackage{graphicx}
\usepackage{microtype}
\usepackage{multirow}
\usepackage{wrapfig}
\usepackage{url}
\usepackage{natbib}
\usepackage{hyperref}
\hypersetup{
  pdftitle={Retracing Hodgkin and Huxley: State Recovery Does Not Certify Mechanism},
  pdfauthor={Peiyu Zang, Jiayi Hao, Yongqiang Cai},
  pdfsubject={cs.LG, q-bio.NC, stat.ML},
  pdfkeywords={Mechanistic Interpretability, Latent State Recovery, Dynamical Systems,
    Scientific Machine Learning, Hodgkin-Huxley Model, Neural Dynamics},
  hidelinks
}

\graphicspath{{figures/}}

\title{Retracing Hodgkin and Huxley:\\State Recovery Does Not Certify Mechanism}

\renewcommand{\shorttitle}{Retracing Hodgkin and Huxley}
\renewcommand{\headeright}{Preprint}
\renewcommand{\undertitle}{Preprint \quad$\cdot$\quad
  \href{https://github.com/factnn/hh-self-discovery}{\textup{\texttt{github.com/factnn/hh-self-discovery}}}}

\author{Peiyu Zang \quad Jiayi Hao \quad Yongqiang Cai\textsuperscript{\ensuremath{\dagger}}\\
School of Mathematical Sciences, Beijing Normal University\\
\texttt{\{202421130105, caiyq.math\}@mail.bnu.edu.cn}\\
\textsuperscript{\ensuremath{\dagger}}Corresponding author}

\date{\today}

\begin{document}

\maketitle

\begin{abstract}
\input{sections/0_abstract}
\end{abstract}


\input{sections/1_introduction}
\input{sections/2_related_work}
\input{sections/3_hh_model}
\input{sections/4_experiments}
\input{sections/5_discussion}
\input{sections/6_conclusion}

\clearpage
\bibliographystyle{preprint}
\bibliography{references}

\input{sections/7_appendix}

\end{document}

%% file: math_commands.tex
\usepackage{amsmath,amsfonts,bm}

\def\eqref#1{equation~\ref{#1}}
\def\1{\bm{1}}

\DeclareMathAlphabet{\mathsfit}{\encodingdefault}{\sfdefault}{m}{sl}
\SetMathAlphabet{\mathsfit}{bold}{\encodingdefault}{\sfdefault}{bx}{n}

%% file: sections/0_abstract.tex
Predicting observed dynamics does not establish recovery of the underlying
physical mechanism. Can machine learning retrace the hidden-state reasoning
behind the Hodgkin--Huxley (HH) model? We train structured latent models on
simulated current and voltage, withholding gate identities and trajectories from
training and model selection. We then test response prediction, state recovery,
protocol transfer, and agreement with HH dynamics. Prediction error and its
cross-seed spread both drop sharply at three latent dimensions under the tested
protocols, while gate recovery under new protocols improves through five to six
coordinates. State recovery depends on which observations the chart uses.
Observed voltage improves current-clamp decoding relative to freely predicted
voltage. Under voltage clamp, adding latent state to command voltage raises
$m$-state $R^2$ from $0.976$ to above $0.99$, yet the transported field
disagrees with HH on identical smooth samples. Known invertible HH coordinates
achieve high fast-$m$ field agreement under the same audit procedure. An exact
HH identity decomposes the discrepancy into time-scale-weighted state error and
a residual in the transported field; these terms can cancel or reinforce. These
findings concern the tested models and charts. They support evaluating state and
dynamics recovery separately, including chart inputs and transported-field
agreement across interventions.

%% file: sections/1_introduction.tex
\section{Introduction}
\label{sec:introduction}

In the late 1940s, Alan Hodgkin and Andrew Huxley confronted a hidden-state
problem. Using voltage clamp on the squid giant axon, they separated an early
inward sodium current from a delayed outward potassium current, but the
mechanism governing these conductances could not be directly observed. From the current
transients they introduced empirical kinetic variables $n,m,h$, with potassium
and sodium conductances represented by $n^4$ and $m^3h$, respectively. These
variables described a proposed mechanism rather than directly observed molecular
gates. The resulting equations reproduced clamp currents, action potentials,
and conduction \citep{hodgkin1952quantitative,huxley2002overshoot}. The
Hodgkin--Huxley (HH) model thus offers a setting for studying scientific
discovery: its responses are observable, its hidden states have physical
meaning, and its governing equations are known. Its history also shows how
controlled experiments can guide the discovery of hidden dynamics.

Neural dynamics span molecular, cellular, and population scales. At the
cellular level, conductance-based models connect ion-channel kinetics to
membrane voltage; modern mechanistic inference and reduced biophysical models
infer cellular dynamics from recorded responses
\citep{gonccalves2020training,clark2022reduced,burghi2025rapid}. At the
population level, latent-state models seek compact dynamical descriptions and,
increasingly, representations that remain meaningful under intervention
\citep{hizli2025identifying,nejatbakhsh2025identifying}. Connecting these scales
raises a basic question: when does a learned representation have a physical
interpretation? We examine this question at the cellular level using HH.
The learner sees only electrical observations, while the experimenter can use
the known hidden states and equations to test what the model has recovered.

% Keep this citation-bearing paragraph together to avoid a link spanning pages.
\begin{samepage}
We revisit the HH discovery as a representation-learning problem. An
experimenter observes current $I(t)$ and membrane voltage $V(t)$, while
the state governing their evolution is hidden. Sequence and latent-dynamics models can
predict future observations from history
\citep{krishnan2015deep,chen2018neural,rubanova2019latent,kidger2020neural},
but prediction alone cannot tell us what they have rediscovered. Can we recover
the physical gating variables from a model's latent state? If so, has the model
also recovered the law governing their evolution?
\par
\end{samepage}

\begin{figure*}[t]
  \centering
  \includegraphics[width=\textwidth]{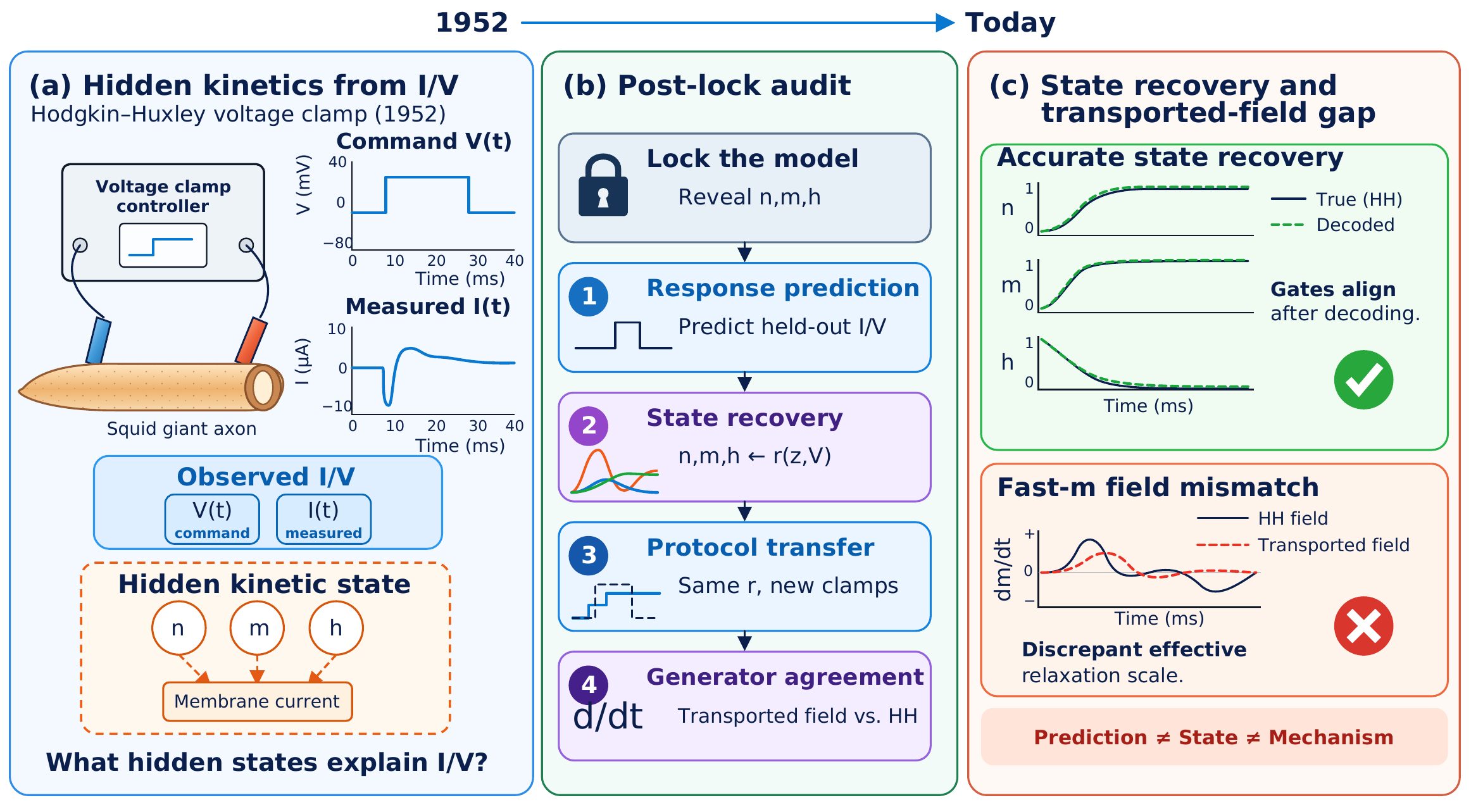}
  \caption{Four tests separate response prediction, state recovery, protocol
  transfer, and generator agreement. Gate information is withheld during training
  and model selection. Accurate gate decoding can coexist with errors in the
  transported fast-$m$ field.}
  \label{fig:overview}
\end{figure*}

Figure~\ref{fig:overview} summarizes our experiment and four tests of
discovery.

These questions require different evidence. Under suitable conditions,
histories of inputs and outputs can distinguish hidden states
\citep{hermann1977nonlinear,takens1981detecting,stark1999delay}. Yet models with
different internal dynamics can predict similar responses under the tested
protocols. A mapping fitted after training, which we call a chart, can decode
$n,m,h$ from the latent state and voltage. Accurate decoded states alone do not
ensure that their time derivatives are correct. To test mechanism recovery,
the same chart must work across interventions, and the dynamics expressed
through it must agree with the HH equations, which we call the generator
\citep{aliee2021beyond,nejatbakhsh2025identifying,zhang2026identifiability}.
We study this gap between recovering physical states and recovering their dynamics.

Existing methods answer important parts of this question.  Data assimilation
infers gates when the conductance model is specified
\citep{toth2011dynamical,moye2018data}; delay models forecast HH
voltage without specifying ion channels \citep{clark2022reduced}; and other
methods learn input--output relations, latent equations, or unknown kinetics
with different amounts of prior structure
\citep{centofanti2024learning,bakarji2023discovering,beck2026learning}.  Recent theory
also distinguishes identifying states from identifying their dynamics and emphasizes
interventions \citep{hizli2025identifying,nejatbakhsh2025identifying}.  What remains unclear
is how to test whether a model recovers both physical states and their dynamics
when those states are not specified in advance.

We test this question using simulated HH trajectories. Models
are trained only on current and voltage histories under current- and
voltage-clamp protocols.  Training, early stopping, and model and dimension
selection never access $n,m,h$. Following the strategy of combining partial scientific structure with
learned dynamics \citep{de2019deep,rackauckas2020universal}, our model assumes
bounded latent states, voltage-dependent relaxation, and a sum of
conductance-like current terms. It is not given gate identities or trajectories,
the $n^4/m^3h$ laws, or the true kinetics. Only after locking each model do we
use simulated gates and HH equations for evaluation. Latent coordinates need not
match physical variables, and accurate state decoding need not imply accurate
derivatives, especially for fast dynamics.\footnote{Code and the reported audit
outputs are available at \url{https://github.com/factnn/hh-self-discovery}.}

Across random seeds, three latent dimensions are the smallest tested setting
whose 20-ms rollout error is reproducible across seeds, and they allow accurate
gate decoding on held-out trajectories
from the training protocol families. More latent coordinates improve gate decoding under new protocols,
although most variance remains in three principal directions. However,
evaluating the same trajectory windows shows that voltage itself contributes
strongly to decoding. Under voltage clamp, charts using the latent state and
command voltage achieve $m$-state $R^2>0.99$, while their transported field agrees
poorly with HH on the identical smooth samples. Sequence and delay models also
contain gate information that transfers across protocols. Equation-discovery
models with and without intervention inputs provide a separate test of whether
learned dynamics respond correctly to those inputs.

An exact HH identity separates the error in the transported field into
time-scale-weighted state error and a residual between those dynamics and the
HH equations at the decoded state. These terms can cancel or reinforce,
explaining how similar state scores can accompany very different errors in
dynamics. Our contributions are to (i) organize hidden-state discovery into
four tests performed after model selection; (ii) identify an empirical $K=3$
transition in prediction error and cross-seed variability, with three-PC
covariance concentration in most overcomplete models and additional coordinates
improving transfer across protocols, without gate supervision; and (iii) show
reproducible gaps between state and dynamics recovery for the tested models and
charts, supported by an exact error decomposition. Retracing Hodgkin and Huxley
returns us to a central question: do the inferred hidden states obey the right
dynamics when the experimental conditions change?

%% file: sections/2_related_work.tex
\section{Related Work}
\label{sec:related-work}

\subsection{From known neuron models to learned hidden states}

Classical approaches start with known conductance equations and infer
hidden channel states and parameters from
current and voltage \citep{vavoulis2012self,meliza2014estimating,
nogaret2016automatic}. Even when the equations are known, whether these
quantities can be uniquely determined depends on the observation protocol and
how the model is parameterized \citep{raue2009structural,villaverde2019full}.
A voltage step may not distinguish steady-state gate values from maximal
conductance. Informative recordings can identify additional kinetics under
assumptions on initial conditions and parameter distinguishability
\citep{csercsik2012identifiability,walch2016parameter}.

Other approaches allow parts of the model to be learned. Data assimilation
can combine observations with ensembles of models to reconstruct neural
dynamics, while hybrid and physics-informed approaches learn effective
currents or unknown kinetics within a specified channel or state structure
\citep{hamilton2014reconstructing,estienne2021towards,
iravanian2020discovery,ghanem2025learning,williams2026correcting}.
We withhold gate identities and trajectories during training and model
selection, then separately test recovery of physical states and their dynamics.

\subsection{From state reconstruction to mechanism recovery}

Delay-coordinate theory describes conditions under which observation histories
can reconstruct states that a single measurement cannot distinguish. These
conditions depend on the dynamics, observation function, and external input
\citep{takens1981detecting,stark1999delay}. Learning methods pursue related
goals: some seek coordinates that simplify the dynamics or organize the state
representation
\citep{lusch2018deep,champion2019data,gilpin2020deep,hill2026geometric};
others reconstruct latent states or discover equations from partial
measurements
\citep{brunton2016discovering,ayed2019learning,ouala2020learning,
lu2022discovering,somacal2022uncovering,bakarji2023discovering,
grigorian2025learning}.

These studies show how to reconstruct hidden information and learn compact
equations. Whether the learned states recover physical variables and their
dynamics under new interventions requires further tests. A chart that decodes
physical states must also work across interventions, and the dynamics expressed
through that chart must agree with the physical generator. We examine these
requirements in HH, where only current and voltage are observed, using
current- and voltage-clamp protocols to test both state recovery and generator
agreement.

%% file: sections/3_hh_model.tex
\section{Hidden-State Discovery in the Hodgkin--Huxley Model}
\label{sec:hh-model}

\subsection{HH dynamics and observation protocols}

We use the classical space-clamped Hodgkin--Huxley (HH) equations as a system
whose observable responses, hidden states, and governing equations are known
\citep{hodgkin1952quantitative}.  With outward current taken as positive, the
current-clamp dynamics are
\begin{align}
C_m\dot V &= I_{\mathrm{app}}-I_{\mathrm{ion}}, \notag\\
I_{\mathrm{ion}} &=
\bar g_{\mathrm{Na}}m^3h(V-E_{\mathrm{Na}})
+\bar g_{\mathrm K}n^4(V-E_{\mathrm K})
+\bar g_{\mathrm L}(V-E_{\mathrm L}), \label{eq:hh-current}\\
\dot x &= \alpha_x(V)(1-x)-\beta_x(V)x
=\frac{x_\infty(V)-x}{\tau_x(V)},
\qquad x\in\{n,m,h\}. \label{eq:hh-gates}
\end{align}
We use the standard parameters
$C_m=1\,\mu\mathrm{F}\,\mathrm{cm}^{-2}$,
$(\bar g_{\mathrm{Na}},\bar g_{\mathrm K},\bar g_{\mathrm L})
=(120,36,0.3)\,\mathrm{mS}\,\mathrm{cm}^{-2}$, and
$(E_{\mathrm{Na}},E_{\mathrm K},E_{\mathrm L})
=(50,-77,-54.4)\,\mathrm{mV}$.  Trajectories are integrated by fourth-order
Runge--Kutta at $\Delta t=0.02\,\mathrm{ms}$, with the gates initialized at
their steady state at the initial voltage.

The same generator is observed under two interventions.  In current clamp,
the command is $I_{\mathrm{app}}(t)$ and the response is $V(t)$.  In voltage
clamp, $V(t)$ is prescribed and the response is the ionic clamp current in
Equation~\ref{eq:hh-current}; we exclude the instantaneous capacitive artifact.
The stimulus family contains current steps, chirps, and piecewise-random
currents, together with voltage steps, prepulse--test--tail sequences, and
additional multistep protocols selected to make responses more sensitive to
changes in the learned latent state.
These modes share hidden state and kinetics but expose them through
different command--response relations.

\subsection{Learning from current and voltage}
\label{sec:observable-formulation}

\begin{wrapfigure}{r}{0.40\textwidth}
  \vspace{-8pt}
  \centering
  \includegraphics[width=\linewidth]{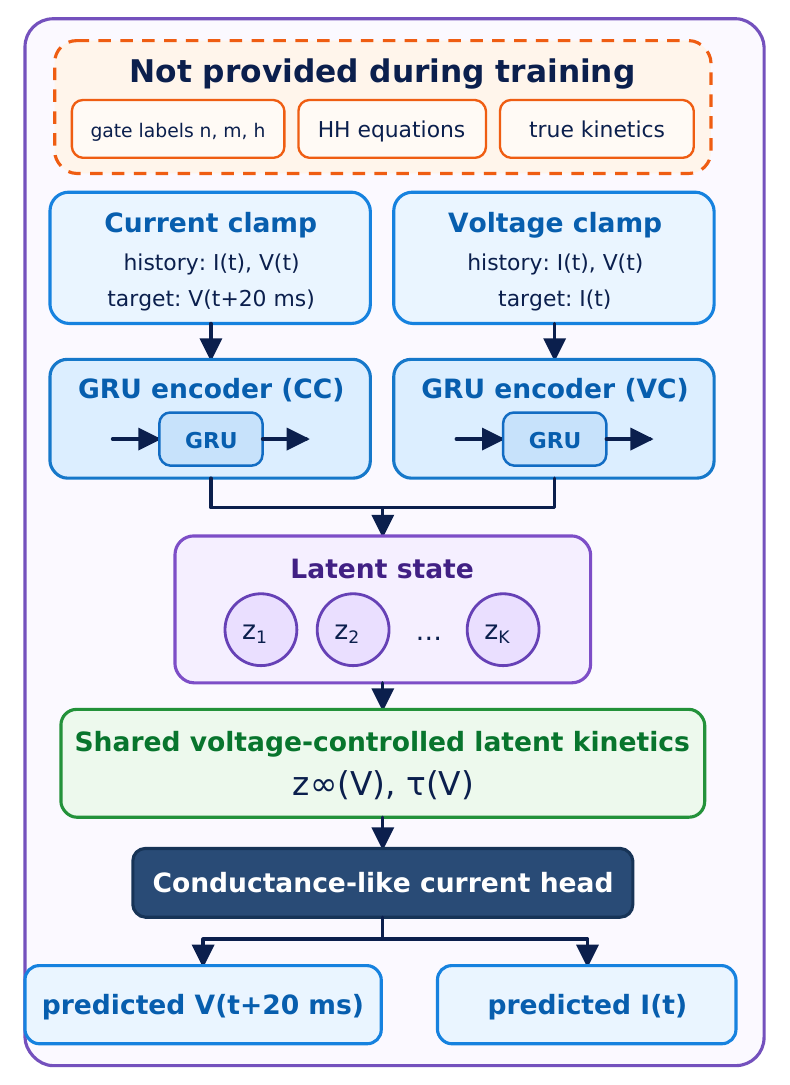}
  \caption{Observable-only predictor with gate labels and HH kinetics withheld.}
  \label{fig:model-architecture}
  \vspace{-8pt}
\end{wrapfigure}

Figure~\ref{fig:model-architecture} summarizes the observable-only predictor and the quantities withheld during training.

For a control mode $c$, let $u_t^c$ denote the command and $y_t^c$ the
response. The model receives an observed history
$\mathcal H_t^c=\{(u_s^c,y_s^c)\}_{s=t-H+1}^{t}$ and a future command, and must
predict the corresponding response over time. During training, validation, early
stopping, and latent-dimension selection, the accessible record contains only
time, current, voltage, and the control mode.  The simulated gates $n,m,h$ and
the separated ionic currents are stored separately for evaluation. An input
check prevents these quantities from entering training or model selection.

The learner is not given gate names, gate trajectories, their number, the
$n^4/m^3h$ powers, or the HH rate functions.  It is nevertheless not a
model without physical assumptions: we assume bounded latent coordinates,
voltage-dependent first-order relaxation, and an additive conductance-like
current law. We ask whether physical states and dynamics can be recovered
within these structural assumptions.

Protocol design uses only fixed, trained models and observable responses. For each
normalized predicted response trajectory
$\hat{\mathbf y}^{\,c}_{1:T}(z_0,\mathbf u)$, we define the latent
output sensitivity and its Gramian:
\begin{equation}
J_c=\frac{\partial \hat{\mathbf y}^{\,c}_{1:T}}{\partial z_0},
\qquad
G_c=\frac{1}{T}J_c^\top J_c,
\qquad
s_c=\lambda_{\min}(G_c).
\label{eq:latent-sensitivity-gramian}
\end{equation}
We estimate $J_c$ by centered finite differences with latent perturbation
$10^{-3}$ and rank candidate commands in descending order by
$\log_{10}\min_j s_c^{(j)} + 0.25\,d_c$, where $d_c$ is the normalized
disagreement between the two trained models' predicted responses. This
empirical protocol score rewards both worst-direction response sensitivity and
model disagreement. No gate or separated-current label enters this calculation.
Because $G_c$ depends on the learned latent coordinates and finite horizon, we
use it as an empirical excitation score, not as a certificate of structural
observability of the HH system.

\subsection{A structured latent model}
\label{sec:latent-realization}

Separate gated recurrent unit (GRU) encoders map each clamp history
$\mathcal H_t^c$ to a latent state $z_t\in(0,1)^K$. Both modes then use the
same latent dynamics. A network maps voltage to equilibrium states and time
constants, and a conductance-like current equation connects the latent state
to the response. There is no separate prediction path that bypasses these
equations. We vary current-branch count $B$ and latent dimension $K$
independently; branches do not represent physical gates.

Training balances samples across trajectories and groups windows by observed
activity and command changes. The loss measures errors in responses and their
step-to-step changes over several prediction horizons. A small regularization
term penalizes latent-state exponents jointly across branches and the summed
branch conductances. We select models by mean normalized validation error
across clamp modes. The dimension scan changes only $K$, and no loss term uses
the true gates.

Holding voltage constant within each step gives the exact latent-state update,
\begin{equation}
z_{t+1}=z_\infty(V_t)+\bigl(z_t-z_\infty(V_t)\bigr)
\exp\!\left[-\Delta t/\tau(V_t)\right].
\label{eq:latent-relaxation}
\end{equation}
The learned ionic current is
\begin{equation}
\widehat I_{\mathrm{ion}}(V,z)=
\sum_{b=1}^{B}g_b\prod_{k=1}^{K}z_k^{p_{bk}}(V-E_b)
+g_{\mathrm L}(V-E_{\mathrm L}),
\qquad g_b,p_{bk}\geq0,
\label{eq:latent-current}
\end{equation}
where conductances, reversal potentials, and exponents are learned.

\subsection{Testing state and dynamics recovery}
\label{sec:discovery-audit}

State recovery relies on observation histories. A single current--voltage pair
need not determine three gates. Observability and delay-reconstruction theory
describe conditions under which input--output histories distinguish hidden
states \citep{hermann1977nonlinear,stark1999delay}. We therefore test whether
the history $\mathcal H_t^c$ supports gate recovery, rather than trying to
invert an instantaneous $(I_t,V_t)$ pair.

After model selection, we fix the model and use the hidden states and HH
equations for evaluation. Four tests proceed from observable responses to
local dynamics. First, we evaluate free response prediction, without supplying
future measured responses. Second, we fit a simple regularized chart
$r:(z,V)\mapsto(\hat n,\hat m,\hat h)$ to decode gates on held-out trajectories
from the training protocol families. Its regularization is selected by
cross-validation with entire trajectories kept together. Third, we evaluate
the fixed chart on unseen protocol families. These tests assess state recovery
and transfer; they do not establish agreement with HH dynamics.

The final test expresses the learned dynamics in gate coordinates using the
chain rule. We call the resulting derivative the transported field:
\begin{equation}
\dot{\hat x}=J_zr(z,V)\dot z+\partial_Vr(z,V)\dot V,
\qquad \hat x=(\hat n,\hat m,\hat h),
\label{eq:pushforward}
\end{equation}
We compare it with the analytic HH field in Equation~\ref{eq:hh-gates},
separately under current and voltage clamp. The full chain rule matters:
accurate decoded gate values do not guarantee accurate derivatives. This
comparison tests generator
agreement for the fitted model--chart pair. Prediction and gate decoding alone
are insufficient evidence.

The HH relaxation law gives an exact decomposition of field error.
Let $e_x=\hat x-x$ and let $\widehat{\dot x}$ denote the transported field
above. Adding and subtracting the HH field at the decoded
state gives
\begin{align}
\widehat{\dot x}-f_x(x,V)
={}&\underbrace{\widehat{\dot x}-f_x(\hat x,V)}_{\text{chart--field residual}}
+\underbrace{f_x(\hat x,V)-f_x(x,V)}_{-e_x/\tau_x(V)} .
\label{eq:field-error-decomposition}
\end{align}
The second term equals $-e_x/\tau_x(V)$ by Equation~\ref{eq:hh-gates}.
State recovery measures chart values on sampled trajectories; dynamics
recovery also depends on chart derivatives through Equation~\ref{eq:pushforward}.
For the same state error, a shorter gate time constant gives a larger second
term. The total field error also depends on the chart--field residual and
whether the two terms cancel or reinforce.

%% file: sections/4_experiments.tex
\section{Experiments}
\label{sec:experiments}

We compare latent dimensions, architectures, and current branches for
response prediction, state recovery, protocol transfer, and generator agreement.

\subsection{Experimental protocol}

The dataset contains 2,600 trajectories generated with fixed HH parameters at
$\Delta t=0.02\,\mathrm{ms}$, split by complete trajectory into 2,080/260/260
train/validation/test examples across three current- and five voltage-clamp
protocol families. Models use a 15-ms history to predict the next 20 ms without
future measured responses; the main dimension study uses five random seeds.
Charts are fitted only after model selection. For tests on unseen protocols
(out-of-distribution, OOD), chart selection uses validation trajectories from
the training protocol families (in-distribution, ID), with entire trajectories
kept in separate folds. The selected charts are then evaluated once on unseen
protocols.

The observable sensitivity--disagreement score selected three additional voltage-clamp
protocols to cover more intervention conditions using only observable data
(Appendix~\ref{app:corpus}).

Independent DOP853 integration, closed-form voltage-clamp solutions, and
step-size refinement verify that simulator error is far below model
discrepancies (Appendix~\ref{app:sim-validation}).

Response normalized root mean squared error (NRMSE) uses the training-response
standard deviation and is
averaged over clamp modes; state recovery is mean $R^2$ over $n,m,h$. Generator
tests use gate-wise transported-field error or, for equation-discovery baselines,
their own held-out field $R^2$. We report means and sample SDs across seeds;
configurations and per-seed results are in the Appendix.

We compare GRU, S4D, a delay-coordinate feedforward neural network (FNN),
Latent ODE, and autonomous and controlled deep delay autoencoders (DDAE).
These baselines test different aspects of recovery and are selected using only
observable validation data
\citep{gu2022parameterization,chen2018neural,rubanova2019latent,bakarji2023discovering}.
Configurations are in Appendix~\ref{app:baselines}.

\subsection{How many latent dimensions are needed?}
\label{sec:dimension-results}

\begin{figure}[!ht]
  \centering
  \includegraphics[width=\textwidth]{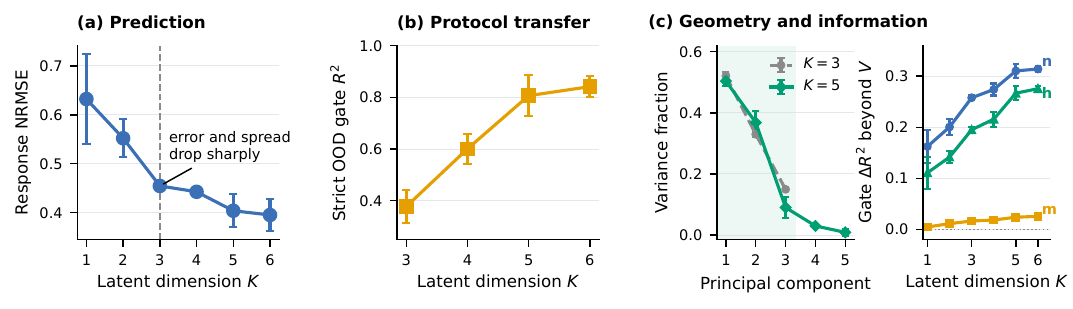}
  \caption{Dimension-dependent diagnostics. (a) Prediction error and cross-seed spread drop at $K=3$; (b) OOD gate recovery plateaus near $K=5$--$6$; (c) latent variance is concentrated in three principal components, while latent state adds mainly $n/h$ information beyond voltage.}
  \label{fig:dimension-transitions}
\end{figure}

Figure~\ref{fig:dimension-transitions} compares prediction, OOD gate recovery, and latent-state variation.

Across five seeds, response NRMSE falls from $0.632\pm0.091$ at $K=1$
to $0.4546\pm0.0004$ at $K=3$, and the cross-seed spread collapses over the
same interval (sample SD $0.039$ at $K=2$ and $0.0004$ at $K=3$). We read $K=3$
as a transition in both the error level and the seed-to-seed reproducibility of
the 20-ms rollout; accuracy continues to improve gradually beyond it, and the
fraction of exactly reproduced spike counts rises from $0.318$ at $K=3$ to
$0.523$ at $K=6$ (Appendix~\ref{app:dimension}). Charts fitted after model selection recover
the ID gates at $K=3$ with mean $R^2=0.9949\pm0.0004$.
Post-lock gate decoding can already be strong below this predictive transition,
underscoring that prediction and state recovery are distinct diagnostics.

OOD gate recovery with charts selected using only ID data increases from $0.378\pm0.065$ at $K=3$ to
$0.806\pm0.079$ at $K=5$ and $0.841\pm0.041$ at $K=6$, indicating a
plateau near $K=5$--$6$. Among the overcomplete models ($K=4$--$6$), at least
95\% of the variance lies in three principal directions for 13 of the 15
trained checkpoints (Table~\ref{tab:dimension-complete}).
Additional coordinates improve OOD gate recovery, while much of their
variation remains predictable from observable history and protocol
information.

A matched comparison of $K=3$ and $K=4$ supports this view: the fourth coordinate
improves chart and field reliability while remaining largely reconstructable
from observable history (Appendix~\ref{app:robustness}).

\subsection{Gate information across models and interventions}
\label{sec:baseline-results}

Generic history models contain substantial transferable gate information, with
S4D providing the strongest predictive and state-decoding control
(Table~\ref{tab:baseline-main}, panel A). This shows that gate recovery is not
specific to our structured architecture. These prediction and decoding tests
do not establish agreement with the HH generator. FNN provides a delay
control (Appendix Figure~\ref{fig:baseline-context}).

DDAE isolates the role of intervention input (Table~\ref{tab:baseline-main},
panel B). Autonomous and controlled variants achieve comparable gate recovery,
but the autonomous model has negative voltage-clamp velocity $R^2$, whereas
the controlled variant retains positive OOD field $R^2$. Thus a model with
sparse latent equations and recoverable gates can still fail when the
command--response relation changes.

The two baseline families answer different questions. The discrete sequence
models (S4D, GRU, FNN) test whether gate information is transferable across
architectures, and their native scores are not defined on HH coordinates. The
Latent ODE admits the same HH-coordinate audit: under one chart class,
support, and metric, it reaches $m$-state
$R^2=0.983\pm0.004$ on held-out in-distribution windows but a transported-field
$R^2$ of $0.068\pm0.089$, against $0.992$ and $0.271$ for the structured model,
and its $n$ and $h$ fields turn strongly negative on unseen protocols
(Appendix Table~\ref{tab:cross-architecture}). A continuous-time latent model
with a different dynamical parameterization therefore exhibits a similar
state--field gap under matched evaluation.

Discrete sequence models enter the same coordinates once their own one-step map
provides a finite-difference estimate of latent velocity. On identical windows
the strongest predictor, S4D, decodes $m$ at $R^2=0.993$ yet reaches only
$0.406$ on the finite-difference transported fast-$m$ field, with a
field-to-truth slope of $0.41$, and a derivative-free exact-step audit shows the
same discrepancy (Appendix Tables~\ref{tab:discrete-flow}
and~\ref{tab:step-flow}). Among the architectures audited in HH coordinates, the
state--field gap remains.

\begin{table*}[t]
\centering
\caption{Diagnostic comparisons with observable-only model selection (mean $\pm$ sample SD; five seeds for structured models, three for baselines). Gate $R^2$ measures ID-to-OOD chart transfer.}
\label{tab:baseline-main}
{\small
\textbf{(A) Predictive representations}\par\smallskip
\resizebox{\textwidth}{!}{%
\begin{tabular}{p{0.25\textwidth}p{0.39\textwidth}rr}
\toprule
Method & Diagnostic role & NRMSE $\downarrow$ & OOD gate $R^2$ $\uparrow$ \tabularnewline
\midrule
Structured ($K=3$) & Three-dimensional latent model & $0.4546\pm0.0004$ & $0.378\pm0.065$ \tabularnewline
Structured ($K=5$) & Additional latent capacity & $0.404\pm0.033$ & $0.806\pm0.079$ \tabularnewline
S4D ($K=5$) & Sequence representation & $\mathbf{0.193\pm0.013}$ & $\mathbf{0.916\pm0.020}$ \tabularnewline
GRU ($K=5$) & Recurrent representation & $0.299\pm0.082$ & $0.841\pm0.118$ \tabularnewline
Latent ODE ($K=2$) & Continuous latent dynamics, command input & $0.510\pm0.060$ & $0.223\pm0.152$ \tabularnewline
\bottomrule
\end{tabular}}

\vspace{5pt}
\textbf{(B) Equation-discovery controls}\par\smallskip
\resizebox{\textwidth}{!}{%
\begin{tabular}{lp{0.27\textwidth}rrr}
\toprule
Method & Diagnostic role & OOD gate $R^2$ $\uparrow$ & VC velocity $R^2$ $\uparrow$ & OOD field $R^2$ $\uparrow$ \tabularnewline
\midrule
DDAE, controlled ($K=3$) & Control-aware dynamics & $0.735\pm0.016$ & $\mathbf{0.628\pm0.101}$ & $0.753\pm0.097$ \tabularnewline
DDAE, controlled ($K=5$) & Control-aware dynamics & $0.619\pm0.016$ & $0.443\pm0.151$ & $\mathbf{0.756\pm0.029}$ \tabularnewline
DDAE, autonomous ($K=3$) & Missing-control test & $\mathbf{0.740\pm0.030}$ & $-0.124\pm0.249$ & $0.391\pm0.039$ \tabularnewline
\bottomrule
\end{tabular}}
}
\end{table*}

\subsection{State recovery does not certify the generator}
\label{sec:fast-gate-results}

% Defer to the next page so that Table 1 (p.~7), Table 2 (p.~8), and
% Figure 4 (p.~9) each sit alone at the top of one page.
\afterpage{%
\begin{table*}[t]
\centering
\caption{Evaluation on shared trajectory windows over five seeds (mean $\pm$ sample SD). Panel A uses complete held-out prediction windows. Panel B evaluates state and transported field on the identical 39,616 smooth voltage-clamp points per model.}
\label{tab:claim-chain-main}
{\scriptsize
\textbf{(A) Observable response and current-clamp conditioning}\par\smallskip
\resizebox{\textwidth}{!}{%
\begin{tabular}{crrrrr}
\toprule
$K$ & CC response NRMSE & $R_m^2(z)$ & $R_m^2(z,V_{\rm obs})$ &
$R_m^2(z,\widehat V)$ & VC response NRMSE \\
\midrule
3 & $.763\pm.003$ & $.045\pm.016$ & $.894\pm.002$ & $.410\pm.008$ & $.151\pm.003$ \\
5 & $.681\pm.077$ & $.466\pm.196$ & $.925\pm.013$ & $.539\pm.122$ & $.139\pm.025$ \\
6 & $.660\pm.049$ & $.527\pm.079$ & $.926\pm.011$ & $.569\pm.060$ & $.135\pm.014$ \\
\bottomrule
\end{tabular}}

\vspace{4pt}
\textbf{(B) State and field evaluated at identical voltage-clamp points}\par\smallskip
\resizebox{\textwidth}{!}{%
\begin{tabular}{crrrr}
\toprule
$K$ & $R_m^2(V_{\rm cmd})$ & $R_m^2(z)$ & $R_m^2(z,V_{\rm cmd})$ &
transported-field $R_m^2$ \\
\midrule
3 & $.9758\pm.0000$ & $.650\pm.031$ & $.9921\pm.0002$ & $.271\pm.006$ \\
5 & $.9758\pm.0000$ & $.826\pm.092$ & $.9944\pm.0012$ & $.303\pm.056$ \\
6 & $.9758\pm.0000$ & $.922\pm.036$ & $.9944\pm.0033$ & $.109\pm.337$ \\
\bottomrule
\end{tabular}}
}
\end{table*}
}

For Table~\ref{tab:claim-chain-main}, each test window starts from a latent state
encoded from observed history. The fixed model then predicts the response
without further measured responses; all charts are
fitted on disjoint validation trajectories. In current clamp (panel A),
conditioning on the observed response voltage substantially improves $m$-state
decoding, but much of this gain disappears when $V_{\rm obs}$ is replaced by
the freely predicted $\widehat V$. Current-clamp chart scores therefore do not
establish equally accurate state recovery during free prediction.

Voltage clamp supplies the cleaner paired test because $V_{\rm cmd}$ is an
external intervention rather than a predicted response. On the identical
smooth voltage-clamp points (panel B), the command alone is already highly predictive
of $m$ ($R^2=0.9758$). The latent state adds measurable information:
$R_m^2(z)$ rises from $0.650$ at $K=3$ to $0.922$ at $K=6$, and
$R_m^2(z,V_{\rm cmd})$ reaches $0.992$--$0.994$, reducing the residual mean
squared error by $67\%$--$77\%$ relative to the command alone. Yet the transported-field
score on those same points is only $0.271$, $0.303$, and $0.109$, respectively.
Thus high state agreement does not certify the fitted model--chart dynamics
(Figure~\ref{fig:state-generator-gap}).

Field agreement does not improve consistently with latent dimension:
$K=6$ remains low and variable despite higher state recovery.
Adding the remaining latent coordinates to $(\hat m,V)$ improves held-out
$R^2$ for predicting the decoded $m$ derivative by $0.15$--$0.16$. Thus, under the tested fit, this derivative still depends on information
outside $(\hat m,V)$. Jumps in
decoded gates at voltage-command changes provide a separate intervention check.
Affine and nonlinear invertible HH controls attain smooth-support $m$-field
$R^2$ of $0.995$ and $0.970$ under the same audit
(Appendix~\ref{app:chart-validity}).
Repeated sampling and initialization with larger neural charts leave the
state--field gap intact (Appendix~\ref{app:temporal-resolution}).

\afterpage{%
\begin{figure}[t]
  \centering
  \includegraphics[width=\textwidth]{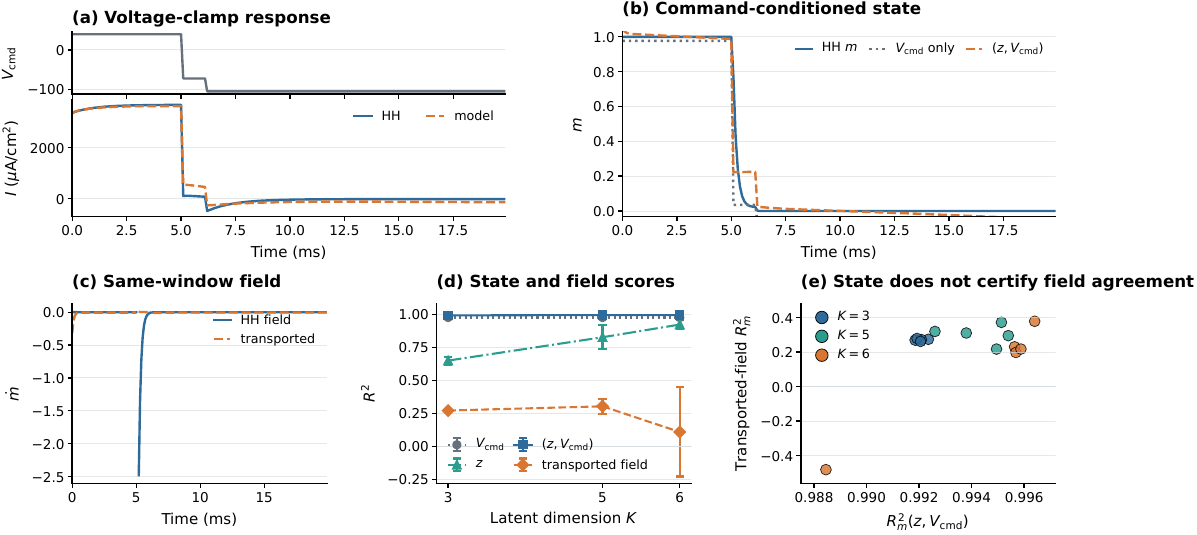}
  \caption{State and transported-field agreement on shared voltage-clamp points. (a--c) $K=5$ example (prespecified distance-to-median rule): command/current, command-conditioned $m$, transported $\dot m$. (d--e) Five seeds show high state $R^2$ but low, variable transported-field $R^2$; bars show sample SD.}
  \label{fig:state-generator-gap}
\end{figure}
}

Equation~\ref{eq:field-error-decomposition} weights state error by
$1/\tau_x(V)$. The larger errors for fast $m$ are consistent with this
sensitivity. We evaluate both this term and the chart--field residual
(Appendix Figure~\ref{fig:mechanism-diagnostics}). A representative chart
gives a slower effective relaxation scale when its transported field is fitted
as a linear function of $m$ at fixed voltage. Because other latent coordinates
still improve derivative prediction, we use this fitted scale as a descriptive
summary of the field mismatch; generator agreement is measured directly by the
transported-field metrics. Across five seeds, two
branch counts, and two chart initializations, $m$-state $R^2$ remains
$0.9992$--$0.9999$ on samples outside voltage-command jump neighborhoods,
despite substantial variation in transported-field error.
We quantify error-term alignment as
$\rho_{\mathrm{align}}
= \langle e_{\mathrm{amp}},e_{\mathrm{chart}}\rangle /
(\lVert e_{\mathrm{amp}}\rVert_2\lVert e_{\mathrm{chart}}\rVert_2)$,
where $e_{\mathrm{amp}}$ is the time-scale-weighted state error
and $e_{\mathrm{chart}}$ the chart--field residual in
Equation~\ref{eq:field-error-decomposition}. Negative and positive values indicate cancellation and reinforcement,
respectively. Thus field error depends on component magnitudes and alignment
even at nearly identical state scores.

Across five checkpoints per branch count, two chart fits each, large-field-error
fits (RMSE $>0.5$) fall from 6/10 at two branches to 0/10 at five, at similar
response and gate recovery. These fits measure chart variability
(Appendix~\ref{app:factorial}).

%% file: sections/5_discussion.tex
\section{Discussion}
\label{sec:discussion}

Our results show a gap between state recovery and dynamics recovery under
the tested protocols. Informative $I/V$ histories can reveal hidden-state
information \citep{hermann1977nonlinear,stark1999delay}, but high gate $R^2$
alone does not show that a chart works across interventions or that its
transported field agrees with the HH generator.

Even with specified HH equations, kinetic identifiability depends on
observation protocols and model assumptions
\citep{csercsik2012identifiability,walch2016parameter}.

Matched retraining at $0.01$-ms resolution preserves the state--field
gap while improving prediction and state recovery
(Appendix~\ref{app:temporal-resolution}).
Equation~\ref{eq:field-error-decomposition} shows how a short time constant
amplifies the state-error term. In the fitted charts, these terms can cancel or
reinforce, producing different field errors at similar state accuracy.

On shared trajectory windows, decoding depends on the supplied voltage.
In current clamp, observed voltage improves decoding over freely predicted
voltage; in voltage clamp, voltage is the imposed command.
The decoded $m$ derivative also retains dependence on other latent coordinates,
and decoded gates jump at voltage-command changes. Known invertible affine and
nonlinear HH coordinates achieve smooth voltage-clamp $m$-field $R^2$ of
$0.995$ and $0.970$ under the same evaluation procedure
(Appendix~\ref{app:chart-validity}). The gap is specific to the tested models
and charts.

These findings suggest designing protocols that help distinguish hidden states
and checking whether the transported field agrees with HH, transfers across
interventions, and preserves gate continuity when commands change. The present
results establish these distinctions in simulated HH dynamics with fixed
parameters and structured latent models. Extending the evaluation to richer
neuronal dynamics and biological recordings is a natural next step.

%% file: sections/6_conclusion.tex
\section{Conclusion}
\label{sec:conclusion}

We asked whether models trained on current and voltage can recover the hidden
states and dynamics of HH. Three findings emerge. First, the prediction error
and its cross-seed spread both fall sharply at three latent dimensions, while
additional coordinates improve gate recovery under new protocols. Second, charts using
latent state and command voltage achieve $m$-state $R^2>0.99$, yet their
transported fields agree poorly with HH on the identical smooth voltage-clamp
samples. Third, an exact error decomposition shows how time-scale-weighted
state error and the chart--field residual can cancel or reinforce. Additional
current branches reduce variation across fitted charts without establishing
dynamics recovery. These conclusions concern the tested models and charts.

For scientific self-discovery, state recovery is only part of the evidence;
studies should also report chart fitting, chart inputs, and transported-field
agreement across interventions.

%% file: sections/7_appendix.tex
\appendix
\section{Appendix}
\label{sec:appendix}

% Organize supplementary material with subsections as the paper develops:
% data generation and protocols, implementation details, complete baseline and
% ablation results, robustness tests, and additional mechanistic audits.

\subsection{Simulator and corpus validation}
\label{app:sim-validation}

The corpus uses the canonical squid-axon HH equations in
Section~\ref{sec:hh-model}.  To avoid circular verification, the reference
implementation duplicates neither the production rate/current routines nor its
fixed-step RK4 integrator.  We rewrote the six rate functions and three ionic
currents independently and integrated them with SciPy's adaptive DOP853 solver
($\mathrm{rtol}=10^{-10}$, $\mathrm{atol}=10^{-12}$).  Table~\ref{tab:sim-validation}
reports one held-out realization of every protocol family.  Current-clamp
voltage error is below $1.54\times10^{-4}\,\mathrm{mV}$; under voltage clamp,
where the commanded voltage is exact, gate RMSE remains below
$1.55\times10^{-3}$.  The larger current differences for the most extreme
active protocols reflect fast gate error amplified by ionic conductance, not a
different command or equation.

\begin{table}[ht]
\centering
\caption{Production RK4 versus independent DOP853.  Response is voltage for
current clamp and ionic current for voltage clamp.}
\label{tab:sim-validation}
\begin{tabular}{lrr}
\toprule
Protocol family & Response RMSE & Gate RMSE \\
\midrule
Current step & $9.60\!\times\!10^{-5}$ mV & $1.74\!\times\!10^{-7}$ \\
Current chirp & $1.52\!\times\!10^{-4}$ mV & $1.48\!\times\!10^{-7}$ \\
Current PRBS & $1.54\!\times\!10^{-4}$ mV & $1.50\!\times\!10^{-7}$ \\
Voltage step & $0.207$ $\mu$A cm$^{-2}$ & $8.17\!\times\!10^{-5}$ \\
Voltage prepulse & $0.064$ $\mu$A cm$^{-2}$ & $7.22\!\times\!10^{-5}$ \\
Active 34 & $0.478$ $\mu$A cm$^{-2}$ & $1.75\!\times\!10^{-4}$ \\
Active 18 & $0.271$ $\mu$A cm$^{-2}$ & $1.54\!\times\!10^{-3}$ \\
Active 09 & $0.148$ $\mu$A cm$^{-2}$ & $9.61\!\times\!10^{-4}$ \\
\bottomrule
\end{tabular}
\end{table}

We add two orthogonal positive controls.  First, at clamps of
$-100,-65,-20,$ and $40\,\mathrm{mV}$, the numerical gates agree with the
closed form $x(t)=x_\infty+(x(0)-x_\infty)e^{-t/\tau_x}$, with maximum absolute
error $2.51\times10^{-5}$.  Second, we rerun every protocol at
$\Delta t=0.02,0.01,0.005\,\mathrm{ms}$ and compare each fixed-step solution
to an adaptive reference driven by the identical sampled command.  Gate error
decreases under refinement for every family.  This establishes numerical
correctness for the simulated canonical HH system, which is a fixed-parameter
squid-axon model; variability across biological preparations is outside the
scope of this study.

\subsection{Corpus construction and hidden-label firewall}
\label{app:corpus}

{\looseness=-1
Each trajectory is independently generated and assigned in its entirety to
train, validation, or test; no history or rollout window crosses a trajectory
boundary.  The fixed corpus contains 2,600 trajectories, split
2,080/260/260.  Current clamp contains step, chirp, and piecewise-random
commands.  Voltage clamp contains ordinary steps, prepulse--test--tail
sequences, and three active multistep families chosen by the observable
sensitivity--disagreement score.  Training files expose only time,
command, response, and clamp mode.  Gates and separated ionic currents are
stored independently and opened only by post-lock audit programs.
\par}

Windows are trajectory balanced and stratified using observable command
transitions, active responses, recovery, and near-steady periods, preventing
quiet samples from dominating the objective.  The standard history is 750
samples ($15\,\mathrm{ms}$), followed by a 1,000-sample
($20\,\mathrm{ms}$) free rollout.  Protocol OOD evaluation withholds complete
stimulus families rather than random windows. Table~\ref{tab:protocols} lists
the families and their sampled parameter ranges.

\begin{table*}[t]
\centering
\caption{Stimulus families.  In-distribution families supply the training,
validation, and test trajectories; out-of-distribution families are withheld
in their entirety and used only for protocol-transfer evaluation.  Current
amplitudes are in $\mu$A$\,$cm$^{-2}$, voltage levels in mV, and each family
reports its trajectory duration.  Parameter ranges are the sampling intervals
used to generate the fixed corpus; CC denotes current clamp and VC voltage
clamp.}
\label{tab:protocols}
\small
\begin{tabular}{>{\RaggedRight\arraybackslash}p{0.15\textwidth}>{\RaggedRight\arraybackslash}p{0.065\textwidth}>{\RaggedRight\arraybackslash}p{0.24\textwidth}>{\RaggedRight\arraybackslash}p{0.40\textwidth}}
\toprule
Family & Mode & Command & Sampled parameters (duration) \\
\midrule
\multicolumn{4}{l}{\emph{In-distribution}}\\
Current step & CC & Step from rest & amplitude $3$--$16$; onset $8$--$20$; offset $55$--$85$ ($100$ ms) \\
Current chirp & CC & Linear frequency sweep & amplitude $3$--$12$; offset $2$--$8$; $0.5$--$5$ to $30$--$100$ Hz ($200$ ms) \\
Current PRBS & CC & Two-level piecewise-random & block $1$--$8$; levels $-4$ and $14$ ($150$ ms) \\
Voltage step & VC & Step with tail & holding $-100$--$-60$; step $-80$--$40$; tail $-100$--$-50$ ($60$ ms) \\
Voltage prepulse & VC & Prepulse--test--tail & holding $-100$--$-70$; prepulse $-120$--$20$; test $-20$--$30$ ($70$ ms) \\
Active multistep ($\times3$) & VC & Four-level multistep & nominal levels chosen by the search (below); levels and durations perturbed during sampling ($50$ ms) \\
\midrule
\multicolumn{4}{l}{\emph{Out-of-distribution}}\\
Current ramp & CC & Triangular ramp & peak $8$--$18$ ($120$ ms) \\
Current multisine & CC & Five superimposed sines & offset $5$; components $0.5$--$2.5$ at $1$--$45$ Hz ($160$ ms) \\
Voltage ramp & VC & Triangular ramp & holding $-100$--$-75$; peak $0$--$50$ ($90$ ms) \\
Voltage tail & VC & Four-level protocol & holding $-100$--$-80$; activate $-10$--$40$; tails $-120$--$-70$ and $-65$--$-30$ ($100$ ms) \\
\bottomrule
\end{tabular}
\end{table*}

The eight in-distribution families supply predictor training (2,080
trajectories), chart fitting and model selection (260 validation
trajectories), and held-out in-distribution evaluation (260 trajectories).
The five parametric families contribute 400 trajectories each and the three
active families 200 each. The four out-of-distribution families contribute 50
trajectories each and are never used for training, for model selection, or for
chart calibration: they form a separate corpus that is evaluated once with the
frozen chart under the strict protocol-transfer protocol of
Appendix~\ref{app:training}.

\paragraph{Observable protocol design.}
For each locked predictor, we compute the finite-horizon latent
output-sensitivity Gramian in
Equation~\ref{eq:latent-sensitivity-gramian}.  The response is normalized by
the checkpoint's training statistics, each encoded latent coordinate is
perturbed by $\pm10^{-3}$ with clipping to $[0,1]$, and Jacobians are pooled
over sampled histories.  The score is coordinate and horizon dependent and is
used only to rank commands for the fixed predictors.  The initial excitation scores have minimum eigenvalues
approximately 2.0--5.9 for current chirp, PRBS, and step families, but only
0.09--0.10 for ordinary voltage steps and prepulses; some voltage protocols
have condition number above 2,000. We searched 60 candidate voltage commands
using only observed histories from the validation split, without reading any
gate values, and confirmed the ranking independently. The
selected active-34, active-18, and active-09 protocols have robust minimum
eigenvalues 0.1648/0.1617/0.1421. Their nominal command levels (holding,
prepulse, test, tail) are $(-108.1,+8.8,+42.2,-76.4)$,
$(-95.1,+11.6,+40.2,-116.1)$, and $(-93.1,-32.7,+20.6,-116.7)$ mV. Each
trajectory spans $50$ ms: $15$ ms at the holding level, the four phases
normalized to occupy the next $20$ ms, and a final $15$ ms at the holding
level. During sampling each level is perturbed by $4$ mV and each phase
duration by $10\%$ before this normalization. Their addition enlarges the
excitation support for subsequent audits.

\subsection{Training, selection, and chart protocol}
\label{app:training}

The locked dimension checkpoints share history length, width-24 kinetics
network, optimization budget, sampler, and observable validation objective.
Five independent seeds, 51--55, form the dimension study.  Later
topology--loss experiments use seeds 56--60 and two independent chart
initializations per checkpoint.  No checkpoint, dimension, branch count, loss
weight, or ridge parameter is selected using test gates.

After locking, a chart maps $(z,V)$ to $(n,m,h)$.  Strict protocol transfer
fits candidate charts on ID trajectories, chooses regularization by
trajectory-grouped five-fold cross-validation with the one-standard-error
rule, and evaluates the frozen chart once on unseen protocol families.  Whole
trajectories form calibration folds.  Repeated windows and initializations are
reliability audits, not opportunities for retrospective model selection.
Absolute transfer scores depend on the checkpoint family and on the ridge the
selection rule picks, so they are comparable only within a fixed setting;
each result below states the family and the selection rule it uses.

Two conventions are used throughout. Field relative RMSE divides the root mean
squared field error by the root mean squared magnitude of the true field on the
same samples. Far-event samples are those more than $2$ ms from any command
transition.

\subsection{Complete latent-dimension results}
\label{app:dimension}

Table~\ref{tab:dimension-complete} separates predictive, alignment, and
geometric dimensions. Prediction improves sharply at $K=3$ and then
gradually. Protocol-invariant alignment continues through $K=5$--$6$, although
nearly every overcomplete checkpoint (13 of 15) remains concentrated around a
three-dimensional covariance core.

\begin{table}[ht]
\centering
\caption{Locked dimension scan. Response NRMSE is reported as the mean over
clamp modes and separately for current clamp (CC) and voltage clamp (VC),
together with the fraction of test windows whose spike count is reproduced
exactly. NRMSE and spike exact are mean $\pm$ sample SD over five seeds.
Protocol transfer is evaluated from $K=3$ onward. The PCA column counts, per
dimension, the checkpoints whose 95\% variance dimension is at most three; the
$K=3$ row holds by construction, so the informative counts are the
$K=4$--$6$ rows, $13$ of $15$ overcomplete checkpoints.}
\label{tab:dimension-complete}
\small
\begin{tabular}{crrrrrr}
\toprule
& \multicolumn{3}{c}{Response NRMSE $\downarrow$} & spike & OOD gate & PCA95 \\
\cmidrule(lr){2-4}
$K$ & mean & CC & VC & exact $\uparrow$ & $R^2$ $\uparrow$ & $\leq3$ \\
\midrule
1 & $0.632\pm0.091$ & $0.886\pm0.067$ & $0.379\pm0.128$ & $0.270\pm0.017$ & -- & -- \\
2 & $0.552\pm0.039$ & $0.846\pm0.033$ & $0.259\pm0.052$ & $0.272\pm0.016$ & -- & -- \\
3 & $0.4546\pm0.0004$ & $0.755\pm0.004$ & $0.154\pm0.004$ & $0.318\pm0.003$ & $0.378\pm0.065$ & 5/5 \\
4 & $0.443\pm0.007$ & $0.747\pm0.012$ & $0.138\pm0.008$ & $0.366\pm0.024$ & $0.601\pm0.057$ & 5/5 \\
5 & $0.404\pm0.033$ & $0.664\pm0.072$ & $0.144\pm0.029$ & $0.490\pm0.106$ & $0.806\pm0.079$ & 4/5 \\
6 & $0.395\pm0.032$ & $0.652\pm0.066$ & $0.139\pm0.017$ & $0.523\pm0.129$ & $0.841\pm0.041$ & 4/5 \\
\bottomrule
\end{tabular}
\end{table}

Two whole-trajectory charts for each $K=3$ checkpoint give ID gate
$R^2=0.9949\pm0.0004$; gate-wise values are
$0.9955/0.9965/0.9926$ for $n/m/h$. However, a matched seed gives mean gate
$R^2=0.937/0.983/0.995$ for $K=1/2/3$: $(z,V)$ can interpolate correlated
on-manifold gates below the transition.  The corresponding
$n/m/h$ field RMSE changes from $0.550/0.974/0.777$ at $K=1$ to
$0.332/0.806/0.468$ at $K=2$ and $0.242/0.721/0.331$ at $K=3$.
Static regression is therefore not used as a dimension certificate alone.

\subsection{Baseline implementation and full comparison}
\label{app:baselines}

The baseline adapters use the official implementations of each method.
Compatibility changes were limited to current Python and pandas imports and to
one correction in the upstream DDAE code. Controlled adapters receive the same
observable histories and commands as the main model; this includes the Latent
ODE baseline, whose original formulation is autonomous and which is therefore
trained with the command as an additional input. FNN remains representation-only.
For DDAE we report both the faithful autonomous model and a declared
controlled extension $\dot z=f(z,u)$.

\begin{figure*}[t]
  \centering
  \includegraphics[width=\textwidth]{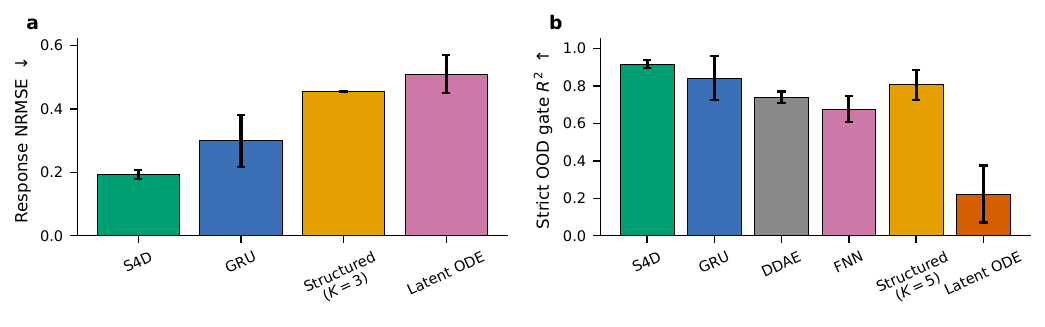}
  \caption{Gate information transfers across tested architectures, while DDAE
  dynamics recovery depends on intervention input. Autonomous DDAE transfers
  under current clamp (CC) but fails under voltage clamp (VC); controlled DDAE
  does not. Error bars show sample SD over five seeds for the structured models
  and three seeds for the baseline families.}
  \label{fig:baseline-context}
\end{figure*}

The commensurate comparisons are reported in main-text
Table~\ref{tab:baseline-main}, which uses five seeds for the structured models
and three seeds for each baseline family;
Figure~\ref{fig:baseline-context} provides the
corresponding visual summary, including the representation-only FNN control.

Autonomous DDAE is the targeted negative control. Its sparse OOD field reaches
$0.786\pm0.073$ in current clamp but $-0.005\pm0.009$ in voltage clamp.
Controlled DDAE restores positive voltage-clamp field performance
($0.631\pm0.152$ at $K=3$ and $0.541\pm0.058$ at $K=5$). Thus sparse,
gate-decodable equations can fail when the intervention changes the
command--response relation.

\subsection{Finite-difference transported-field audit}
\label{app:discrete-flow}

A discrete sequence model has no latent velocity, so it cannot enter the
transported-field audit directly. It does define a one-step map, and we
estimate the velocity from that map, $\dot z\approx(z_{t+1}-z_t)/\Delta t$,
along the model's own free rollout. The quantity compared with the HH field is
therefore a finite-difference transported field, not an instantaneous flow, and
we label it as such throughout. The derivative-free comparison reported at the
end of this subsection does not rely on that difference. Everything else is held fixed: the same
cubic ridge chart with the same coefficient, the same in-distribution
validation windows for fitting, the same held-out smooth voltage-clamp support,
and the same chain rule and metric. On the structured model this
finite-difference velocity reproduces the analytic-velocity field score
($0.899/0.269/0.743$ against $0.897/0.271/0.740$ for $n/m/h$), so the two
constructions agree on the samples used here.

Table~\ref{tab:discrete-flow} applies that audit to four architectures on
identical windows; the decoded $n$ and $h$ states are $0.96$--$0.99$ except for
the Latent ODE ($0.85$ and $0.91$). Every model decodes $m$ at $R^2\ge0.98$, and the strongest
predictor of the set, S4D, also has the best $m$ field score. It still reaches
only $0.406$, and the slope of the transported field against the true field is
$0.41$: the fast-$m$ field has a substantially reduced field-to-truth slope,
with correlation $0.64$. S4D's $n$ and $h$ fields are comparatively strong ($0.811$ and
$0.849$), so the residual failure is again specific to the fast gate rather
than to the architecture. GRU and the Latent ODE sit below the structured model
on the same metric, with $m$ slopes of $0.17$ and $0.04$. Among the
architectures evaluated with the common HH-coordinate audit, the state--field
gap remains.

\begin{table*}[t]
\centering
\caption{Finite-difference transported-field audit on identical held-out smooth voltage-clamp windows
(39{,}032 consecutive pairs on the shared 0.1-ms evaluation grid, that is,
every fifth sample of the native 0.02-ms stream; the unpaired smooth support is
39{,}616 samples). Each model's own finite-difference latent velocity over that
interval is transported through the same cubic chart and compared with the
canonical HH gate field, on held-out in-distribution windows and on unseen
protocol families. State and field columns are $R^2$ (higher is better);
``slope'' is the regression coefficient of the transported $m$ field on the true
$m$ field, so exact agreement implies a slope of $1$. Pairs do not cross command
transitions. Values are mean $\pm$ sample SD over five structured seeds
or three seeds per baseline. The Latent ODE has no unseen-protocol row here
because its transfer scores are reported in Table~\ref{tab:cross-architecture}.}
\label{tab:discrete-flow}
\small
\resizebox{\textwidth}{!}{%
\begin{tabular}{lccccc}
\toprule
Model & state $m$ & field $n$ & field $m$ & field $h$ & $m$ slope \\
\midrule
Structured ($K=3$) & $\mathbf{0.992\pm0.000}$ & $0.899\pm0.008$ & $0.269\pm0.006$ & $0.743\pm0.024$ & $0.220\pm0.009$ \\
S4D ($K=5$) & $0.993\pm0.001$ & $0.811\pm0.120$ & $\mathbf{0.406\pm0.045}$ & $0.849\pm0.074$ & $0.414\pm0.080$ \\
GRU ($K=5$) & $0.991\pm0.003$ & $0.564\pm0.219$ & $0.205\pm0.176$ & $0.659\pm0.173$ & $0.165\pm0.136$ \\
Latent ODE ($K=2$) & $0.983\pm0.004$ & $0.023\pm0.335$ & $0.060\pm0.091$ & $0.387\pm0.048$ & $0.038\pm0.054$ \\
\midrule
\multicolumn{6}{l}{\emph{Unseen protocol families, chart transferred from in-distribution validation}}\\
Structured ($K=3$) & $0.980\pm0.001$ & $0.664\pm0.039$ & $0.058\pm0.010$ & $0.530\pm0.044$ & $0.046\pm0.006$ \\
S4D ($K=5$) & $0.974\pm0.004$ & $0.296\pm0.360$ & $0.195\pm0.115$ & $0.622\pm0.144$ & $0.191\pm0.062$ \\
GRU ($K=5$) & $0.965\pm0.009$ & $0.114\pm0.465$ & $0.114\pm0.153$ & $0.291\pm0.405$ & $0.091\pm0.104$ \\
\bottomrule
\end{tabular}}
\end{table*}

Under protocol transfer the same picture holds with a wider margin. The chart is
fitted on in-distribution validation windows only and evaluated once on unseen
protocol families, as in Table~\ref{tab:cross-architecture}. Structured, S4D,
and GRU keep high decoded $m$ state ($R^2=0.965$--$0.980$), while the transported
fast-$m$ field falls to $0.058$, $0.195$, and $0.114$ with slopes of $0.05$,
$0.19$, and $0.09$. The strongest predictor loses most of its in-distribution
field score, and the fast-$m$ field retains a substantially reduced
field-to-truth slope in all three models.

Finite-difference transport does not generally equal the instantaneous
transported field for a discrete sequence model. We therefore add a
derivative-free one-step comparison. From the same decoded state $r(z_t,V)$ we compare the model's own
next decoded state $r(z_{t+1},V)$ with the exact HH relaxation step
$x_\infty(V)+[r(z_t,V)-x_\infty(V)]e^{-\Delta t/\tau(V)}$, which is exact under
voltage clamp because the command is constant over a step
(Table~\ref{tab:step-flow}).

The absolute next state is uninformative here, because one step changes a gate
very little and every model reaches $R^2\ge0.998$ against the reference. The
increment is what discriminates. For the fast gate, the model's own next step
still disagrees strongly with the HH next step from the identical starting
point: $R^2=0.140\pm0.010$ for the structured model, $0.335\pm0.058$ for S4D,
$0.174\pm0.122$ for GRU, and $-0.034\pm0.097$ for the Latent ODE, with slopes
between $0.02$ and $0.27$, while the trivial predictor that the gate does not
change scores about zero. The $n$ and $h$ increments agree much better for the
structured model, S4D, and GRU ($0.64$--$0.91$ and $0.72$--$0.87$), although the
Latent ODE recovers the $n$ increment only weakly ($0.158\pm0.673$) and its $h$
increment is also lower ($0.565\pm0.050$). The
cross-architecture conclusion therefore does not depend on the derivative
approximation.

\begin{table}[ht]
\centering
\caption{Derivative-free one-step flow on the same smooth voltage-clamp samples.
Each model's own next decoded state, one 0.1-ms step later, is compared with
the exact HH relaxation step from the identical decoded state over the same
interval; no Jacobian or instantaneous derivative is involved. Values are increment (change) $R^2$, mean $\pm$ sample SD over five
structured seeds or three seeds per baseline; ``persistence'' is the trivial
predictor that the gate does not change.}
\label{tab:step-flow}
\small
\setlength{\tabcolsep}{5pt}
\begin{tabular}{lcccc}
\toprule
Model & $n$ increment & $m$ increment & $h$ increment & $m$ slope $\to 1$ \\
\midrule
Structured ($K=3$) & $0.911\pm0.006$ & $0.140\pm0.010$ & $0.730\pm0.019$ & $0.106\pm0.007$ \\
S4D ($K=5$) & $0.850\pm0.098$ & $\mathbf{0.335\pm0.058}$ & $0.865\pm0.054$ & $0.267\pm0.068$ \\
GRU ($K=5$) & $0.640\pm0.177$ & $0.174\pm0.122$ & $0.720\pm0.110$ & $0.138\pm0.077$ \\
Latent ODE ($K=2$) & $0.158\pm0.673$ & $-0.034\pm0.097$ & $0.565\pm0.050$ & $0.019\pm0.031$ \\
\midrule
Persistence (all models) & $-0.02$ to $0.00$ & $-0.07$ to $0.00$ & $-0.02$ to $0.00$ & -- \\
\bottomrule
\end{tabular}
\end{table}

Decoding gates at consecutive samples is a different question, because it also
contains the chart's nonlinearity over the interval. For fast $m$, a finite step over the evaluation
interval need not coincide with the instantaneous derivative. We therefore use the derivative-level audit and the derivative-free
exact-step comparison as complementary diagnostics, and report the decoded
increment as a secondary check. Repeating the
field comparison at the native 0.02-ms interval also yields low fast-$m$ field
scores: $0.081$, $0.339$, $0.162$, and $0.017$ for the structured model, S4D,
GRU, and Latent ODE, respectively.

\subsection{Chart reliability and regularization}
\label{app:chart-reliability}

Whole-trajectory grouped-CV calibration substantially improves test--retest
stability and reproduces the $K=3$--$6$ ordering across repeated windows and
fold assignments.

The one-standard-error grouped-CV rule is repeated over independent windows
and fold assignments; it also recovers a $K=5$ checkpoint that point-level
calibration does not.

Whole-trajectory holdout reduces test--retest variation. For a $K=4$
checkpoint, point-calibrated total $m$-field RMSE spans 0.602--0.847
(SD 0.125), whereas two trajectory-calibrated fits give 0.669/0.709
(SD 0.028). For a cleaner $K=6$ checkpoint, SD falls from 0.056 to 0.006.
Across all five $K=4$ checkpoints, two trajectory-level fits attain
test--retest Spearman 0.90, Pearson 0.968, ICC 0.671, and mean absolute
difference 0.035 for total $m$ error.

On the same support the latent Jacobian $\partial r/\partial z$ has median
condition number $11$--$23$ at $K=3$, with a median smallest singular value of
$0.10$--$0.17$ and a 5th percentile of $0.01$--$0.02$, and $8$--$10$ at
$K=5$--$6$, and the pointwise
$m$-field error is essentially uncorrelated with the local condition number
(Spearman $-0.17$ to $+0.16$ across checkpoints, and $+0.13$ to $+0.52$ against
the smallest singular value). The examined local conditioning metrics do not
therefore explain the observed fast-$m$ field error.

A dynamics-aware chart objective makes the state--field trade explicit instead
of leaving it to the selection rule. We sweep the weight $w$ on the transported
field term in the chart objective, holding the checkpoint, the cubic chart
class, and the validation split fixed. Table~\ref{tab:chart-pareto} shows the
result for four locked checkpoints: increasing $w$ from $0$ to $1$ reduces
field relative RMSE from $1.31$--$4.78$ to $0.93$--$0.97$ while lowering decoded
gate $R^2$ by $4$--$8$ percentage points. Increasing the field-loss weight generally reduces field error at the
cost of state accuracy. None of the tested settings achieves low errors on both
quantities. Three further chart initializations and training budgets extend the
sweep to $w=10$, where the gate $R^2$ falls to $0.66$--$0.86$ and the best
field relative RMSE over all runs is still $0.73$. The trade is therefore an
empirical property of the tested fits rather than of one selection rule.

\begin{table}[ht]
\centering
\caption{Dynamics-aware chart sweep on the same validation split and evaluation
windows. Cells are decoded gate $R^2$ / transported-field relative RMSE; lower
field error is better. Increasing the field weight traces an empirical
state--field trade-off and never reaches a low-error field.}
\label{tab:chart-pareto}
\small
\setlength{\tabcolsep}{5pt}
\begin{tabular}{lcccc}
\toprule
Checkpoint & $w=0$ & $w=0.01$ & $w=0.1$ & $w=1$ \\
\midrule
$K=5$, seed 51 & $0.909$ / $1.31$ & $0.909$ / $1.11$ & $0.900$ / $1.00$ & $0.869$ / $0.96$ \\
$K=5$, seed 54 & $0.904$ / $3.25$ & $0.897$ / $1.21$ & $0.879$ / $1.01$ & $0.822$ / $0.97$ \\
$K=6$, seed 52 & $0.926$ / $2.51$ & $0.925$ / $1.25$ & $0.912$ / $0.97$ & $0.871$ / $0.93$ \\
$K=6$, seed 55 & $0.889$ / $4.78$ & $0.871$ / $1.23$ & $0.859$ / $1.01$ & $0.807$ / $0.97$ \\
\bottomrule
\end{tabular}
\end{table}

\subsection{State recovery and the fast-$m$ field gap}
\label{app:mechanism}

The five $K=3$ checkpoints give Table~\ref{tab:k3-field}. The fast-$m$ field
discrepancy is consistent across them: total RMSE is $0.713\pm0.009$, with
state-amplification and chart--field residual components
$0.678\pm0.010$ and $0.213\pm0.005$.

\begin{table}[ht]
\centering
\caption{Minimal $K=3$ physical audit over five checkpoints.}
\label{tab:k3-field}
\begin{tabular}{lrrr}
\toprule
Gate & $n$ & $m$ & $h$ \\
\midrule
State $R^2$ & 0.9955 & 0.9965 & 0.9926 \\
Field relative RMSE & 0.322 & 0.713 & 0.373 \\
\bottomrule
\end{tabular}
\end{table}

We stratify the post-lock audit by the true HH time constant $\tau_m(V)$.
Splitting the same smooth voltage-clamp samples into quartiles of
$\tau_m(V)$ (Table~\ref{tab:tau-stratification}), the transported $m$ field is
essentially uncorrelated with the true field wherever $\tau_m<0.16$ ms
($\rho=0.03$--$0.06$, slope $\approx0$) and only partially correlated above it
($\rho\approx0.75$), where the field-to-truth slope is still only
$0.35$--$0.47$. The decoded $m$ *state* is equally accurate in every quartile,
so the difference is in the dynamics rather than in the decode. Stratifying by
the magnitude of the true $m$ derivative is uninformative: that derivative is
near zero on most samples, so relative error is dominated by the denominator.

\begin{table}[ht]
\centering
\caption{Transported $m$ field by HH time-constant quartile on the shared smooth
voltage-clamp samples, mean over the five $K=3$ checkpoints.}
\label{tab:tau-stratification}
\small
\setlength{\tabcolsep}{5pt}
\begin{tabular}{lrrrr}
\toprule
$\tau_m(V)$ quartile & Q1 & Q2 & Q3 & Q4 \\
\midrule
Range (ms) & $0.008$--$0.054$ & $0.054$--$0.158$ & $0.158$--$0.201$ & $0.201$--$0.501$ \\
State $R^2_m$ & $0.992$ & $0.992$ & $0.991$ & $0.992$ \\
Field relative RMSE & $1.00$ & $1.00$ & $0.66$ & $0.72$ \\
Field correlation & $0.03$ & $0.06$ & $0.76$ & $0.74$ \\
Field slope & $0.00$ & $0.01$ & $0.47$ & $0.35$ \\
\bottomrule
\end{tabular}
\end{table}

\begin{figure*}[t]
  \centering
  \includegraphics[width=\textwidth]{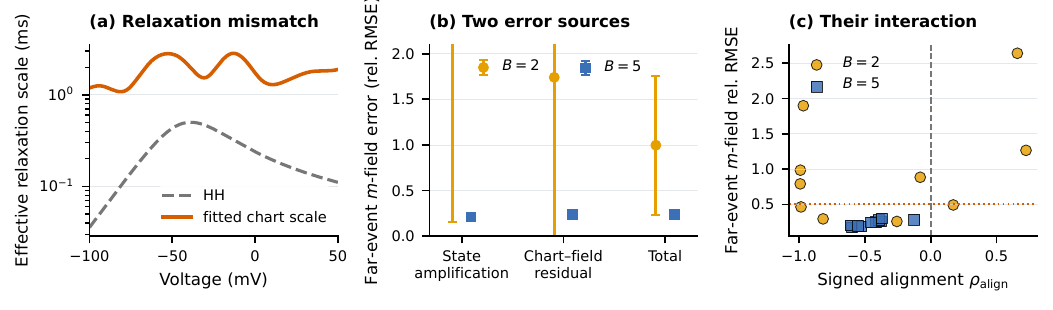}
  \caption{Fast-$m$ field diagnostics: (a) fitted effective-relaxation mismatch; (b--c) state-amplification and chart--field residual terms whose alignment determines cancellation or reinforcement.}
  \label{fig:mechanism-diagnostics}
\end{figure*}

Decoded fields admit low-complexity affine-in-gate relaxation fits over the
empirical support. Steady-state ordering and slow $n/h$ scale structure are
stable across $K=5$--$6$ checkpoints. For $m$, the fitted steady state remains
informative but the effective relaxation scale is systematically several-fold
slower than HH across the audited operating points. Because closure is not
established, this scale is descriptive: none of the tested shared rescalings,
gate-specific constants, or low-degree voltage warps restores the analytic
$m$ field across the audited checkpoints.

Numerical controls validate this conclusion. Analytic HH velocities agree
with finite differences on true current-clamp trajectories with RMSE 0.0455
and cosine 0.99992. Learned analytic velocities agree with finite differences
of their own trajectories with cosine 0.997--0.9997. Removing command-jump
neighborhoods and adding voltage to the chart with the full chain rule do not
remove the fast-$m$ discrepancy.

\subsubsection{Learner temporal-resolution control}
\label{app:temporal-resolution}

We regenerated all 2,600 trajectories with identical protocol draws and
trajectory splits at $\Delta t=0.01\,\mathrm{ms}$, doubled the history and
rollout lengths to preserve the physical durations of 15 and 20 ms, and
retrained the $K=5/6$ models for three paired seeds. The $0.02$-ms checkpoints
were re-audited using the same windows and grouped-CV chart procedure. Finer
sampling improves prediction and state recovery, while the state--field
gap persists (Table~\ref{tab:sampling-rate}).

\begin{table}[ht]
\centering
\caption{Matched learner sampling-rate control (mean over three paired seeds).
The $m$-state $R^2$ is evaluated over all points in event-balanced ID test
windows using the grouped-CV cubic post-lock chart; it is not the far-event
subset or dynamics-aware chart score reported in Figure~\ref{fig:mechanism-diagnostics}.}
\label{tab:sampling-rate}
\small
\begin{tabular}{ccrrrr}
\toprule
$K$ & $\Delta t$ (ms) & NRMSE $\downarrow$ & $m$ state $R^2$ $\uparrow$ &
$\tau_m^{\mathrm{eff}}/\tau_m^{\mathrm{HH}}$ $\downarrow$ & $m$-field $R^2$ $\uparrow$ \\
\midrule
5 & 0.02 & 0.429 & 0.768 & 14.44 & 0.030 \\
5 & 0.01 & 0.389 & 0.882 & 13.32 & 0.046 \\
6 & 0.02 & 0.418 & 0.874 & 10.16 & 0.054 \\
6 & 0.01 & 0.326 & 0.939 & 8.98 & 0.094 \\
\bottomrule
\end{tabular}
\end{table}

The design choices in Table~\ref{tab:exclusion-chain} also cover chart capacity and objective choice.
Voltage-conditioned charts with the complete chain rule improve mean gate
$R^2$ from roughly 0.916 to 0.981 and field RMSE from 0.914 to 0.736,
showing that omitted voltage explains part of the early conflict.
Clamp-specific charts improve current-clamp RMSE to 0.164--0.216, but
voltage-clamp RMSE remains 0.72--0.76. A four-layer, 256-unit, 20,000-step
chart improves selected fits, but repeated locked resampling does not
consistently remove the state--field gap.

\subsubsection{Chart closure, command jumps, and positive control}
\label{app:chart-validity}

We re-audited 15 locked $K=3/5/6$ checkpoints on exactly shared prediction,
state, and field windows. Each window uses one observed-history encoding
followed by a free latent rollout. The 200 windows per clamp mode are drawn by
selecting held-out test trajectories uniformly and then a window uniformly
within each, so the smooth voltage-clamp support pools $39{,}616$ samples from
$94$ distinct test trajectories across the five voltage-clamp families, with
every window weighted equally regardless of trajectory length. Under current
clamp (CC), a cubic chart
receives $z$, $(z,V_{\rm obs})$, or $(z,\widehat V)$, where $\widehat V$ is the
freely predicted voltage response. Under voltage clamp (VC), voltage is the
externally imposed command: both voltage-conditioned branches use the identical
$V_{\rm cmd}$ and $\dot V_{\rm cmd}$, so no predicted-voltage comparison exists.
Table~\ref{tab:claim-chain} therefore reports the two modes separately.

\begin{table}[ht]
\centering
\caption{Strict same-support VC audit (five-seed mean $\pm$ sample SD).
All state and field scores use the identical 39,616 held-out samples satisfying
$|\dot V_{\rm cmd}|\leq1\,\mathrm{mV/ms}$ per checkpoint.  Because the command
is piecewise constant on the sampling grid, this threshold selects exactly the
samples on which the command is constant between level changes, so the
transported field there is the chart pushforward with the command frozen.}
\label{tab:claim-chain}
\small
\begin{tabular}{crrrrr}
\toprule
$K$ & $R_m^2(V_{\rm cmd})$ & $R_m^2(z)$ & $R_m^2(z,V_{\rm cmd})$ &
field $R_m^2$ & field median \\
\midrule
3 & $.9758\pm.0000$ & $.650\pm.031$ & $.9921\pm.0002$ & $.271\pm.006$ & .273 \\
5 & $.9758\pm.0000$ & $.826\pm.092$ & $.9944\pm.0012$ & $.303\pm.056$ & .311 \\
6 & $.9758\pm.0000$ & $.922\pm.036$ & $.9944\pm.0033$ & $.109\pm.337$ & .218 \\
\bottomrule
\end{tabular}
\end{table}

The command alone explains much of the smooth-support pointwise $m$ variation,
while the latent state adds measurable information. Full-window CC results are
reported in Table~\ref{tab:claim-chain-main}; the corresponding $m$-field $R^2$ under
$V_{\rm obs}/\widehat V$ are $.665/.130$, $-.749/.209$, and
$-10.664/.145$ for $K=3/5/6$. The apparent improvements under $\widehat V$ for
$K=5/6$ reflect compensation among voltage, state, and chart errors when
evaluated jointly off the fitted trajectory.

Closure is tested on held-out trajectories by predicting the transported
$\dot{\hat m}$ nonparametrically from $(\hat m,V)$, then from
$(\hat m,V,z)$. Restoring $z$ raises $R^2$ by 0.15--0.16, providing evidence
that $(\hat m,V)$ is insufficient for the tested transported field. Two
controls show that the increment measures the coordinate rather than the
regressor. Across five regressor settings (tree counts, leaf sizes, and a
gradient-boosting alternative) the learned increment stays between $+0.13$ and
$+0.18$; and in the known-coordinate controls, where the decoded $\hat m$ is
correct by construction, the same comparison returns $+0.004$ and $+0.024$ when
the $(\hat m,V)$ baseline is well estimated. The control increment grows only
when that baseline is itself underfit (relative $R^2$ $0.40$--$0.71$ instead of
$0.89$--$0.92$). These controls therefore support additional latent-state
dependence beyond the tested regressor-capacity effects. At voltage-command
jumps, holding $z$ fixed while changing only $V$ produces median chart jumps of
0.21--0.25 in $m$ (95th percentile 0.94--0.99), versus a median true one-step
change of 0.027. The two quantities are not the same object: the chart jump
measures the continuity of the fitted map $r(z,V)$ across a command
discontinuity, whereas the one-step change includes the physical relaxation of
the gate over that interval. We report both for that reason, and we separate
smooth-support field evidence from jump artifacts and treat fitted
$\tau_m^{\mathrm{eff}}$ as descriptive.

As an end-to-end positive control, we hide the exact HH gates behind known
invertible affine and smooth nonlinear coordinates, refit the same cubic chart class,
and apply the same JVP transport over three independent window samples. On
smooth voltage-clamp support, affine-coordinate field $R^2$ is
$0.974/0.995/0.989$ for $n/m/h$; the nonlinear control gives
$0.747/0.970/0.887$. On smooth voltage-clamp support the known-coordinate
controls therefore recover high field agreement, showing that the audit detects
known smooth conjugacies; command-jump neighborhoods are analyzed separately.
Specifically, for $x=(n,m,h)^\top$ we use
\begin{align*}
z_{\rm aff}&=A_{\rm aff}x+b_{\rm aff}, &
A_{\rm aff}&=\begin{bmatrix}.90&.18&-.12\\-.14&1.05&.16\\.10&-.20&.94\end{bmatrix}, &
b_{\rm aff}&=(.08,-.05,.04)^\top,\\
z_{\rm nl}&=\sigma\!\left(A_{\rm nl}\operatorname{logit}(x)+b_{\rm nl}\right), &
A_{\rm nl}&=\begin{bmatrix}1&.16&-.10\\-.12&.92&.14\\.08&-.15&1.06\end{bmatrix}, &
b_{\rm nl}&=(.10,-.08,.06)^\top.
\end{align*}
Both matrices are nonsingular. Replicates use seeds 19700--19702, 300 validation
and 300 test windows per clamp mode, stride 5, ridge coefficient 100, and a
$0.01\,\mathrm{ms}$ centered JVP step. The affine inverse lies in the cubic
chart class; its nonzero error therefore reflects finite sampling,
regularization, and numerical JVP approximation rather than missing capacity.

\begin{table}[ht]
\centering
\caption{State and transported-field recovery for different charts, evaluated
on the same smooth voltage-clamp support ($|\dot V_{\rm cmd}|\le1$ mV/ms). The
learned rows use the shared $39{,}616$-point support, one free rollout per
window, on locked checkpoints; the control rows replace the model with the exact
HH gates behind a known invertible conjugacy, resample smooth-support windows
independently under the same criterion ($59{,}426$ points over three
replicates), and refit the same cubic chart class.  Values are mean $\pm$ sample SD over five checkpoints or three
control replicates.}
\label{tab:chart-classes}
\small
\begin{tabular}{lrrr}
\toprule
Chart & $m$-state $R^2$ $\uparrow$ & $m$-field $R^2$ $\uparrow$ & $n$ \\
\midrule
Linear, $K=3$ & $0.941\pm0.002$ & $0.039\pm0.004$ & 5 \\
Cubic, $K=3$ & $0.9921\pm0.0002$ & $0.271\pm0.006$ & 5 \\
Linear, $K=5$ & $0.942\pm0.003$ & $0.157\pm0.073$ & 5 \\
Cubic, $K=5$ & $0.9944\pm0.0012$ & $0.303\pm0.056$ & 5 \\
\midrule
Known affine HH & $1.0000\pm0.0000$ & $0.995\pm0.001$ & 3 \\
Known nonlinear HH & $0.9999\pm0.0000$ & $0.970\pm0.010$ & 3 \\
\bottomrule
\end{tabular}
\end{table}

The gap is not unique to the cubic chart: a linear chart, which
fits the gates far less well ($m$-state $R^2$ $0.941$ against $0.992$), shows
the gap in the same direction and with a wider margin ($m$-field $R^2$ $0.039$
against $0.271$ at $K=3$). Both scores come from the same windows and the same
support, so the comparison isolates the chart rather than the evaluation.

\begin{table}[ht]
\centering
\caption{Cross-architecture state and transported-field recovery using the same
cubic chart class, ridge coefficient, smooth voltage-clamp support, and
evaluation metric.  The
Latent ODE is a continuous-time latent flow with a parameterization unrelated
to the structured model and comes from a different training batch; the state
and field columns are matched window by window.  Values are mean
$\pm$ sample SD over five structured checkpoints or three Latent ODE seeds.}
\label{tab:cross-architecture}
\small
\setlength{\tabcolsep}{4pt}
\begin{tabular}{lrrrr}
\toprule
Model, split & $R^2_m(z,V_{\rm cmd})$ & field $R^2_m$ & field $R^2_n$ & field $R^2_h$ \\
\midrule
Structured, ID test & $0.9921\pm0.0002$ & $0.271\pm0.006$ & $0.896\pm0.009$ & $0.739\pm0.024$ \\
Latent ODE, ID test & $0.983\pm0.004$ & $0.068\pm0.089$ & $-0.018\pm0.368$ & $0.338\pm0.084$ \\
Structured, unseen & $0.980\pm0.001$ & $0.057\pm0.010$ & $0.665\pm0.040$ & $0.531\pm0.045$ \\
Latent ODE, unseen & $0.950\pm0.017$ & $0.010\pm0.053$ & $-3.57\pm4.91$ & $-1.93\pm2.76$ \\
\bottomrule
\end{tabular}
\end{table}

The Latent ODE reproduces the state--field gap with a different latent
geometry. Despite accurate $m$-state decoding, its transported field shows low
agreement with HH on the same smooth voltage-clamp support, and on unseen
protocols its $n$ and $h$ field scores are strongly negative, unlike those of
the structured model.

\subsection{Architecture and transient-loss factorial study}
\label{app:factorial}

Branch count $B$ controls current-readout capacity, not latent dimension.
Table~\ref{tab:branch-scan} rules out one branch per gate: two- and
five-branch models recover all three gates. Across three initial seeds,
NRMSE for $B=2/3/5$ is $0.418/0.429/0.404$, so the single-seed three-branch
prediction optimum does not persist.

\begin{table}[ht]
\centering
\caption{Branch-count control at $K=5$, seed 56.}
\label{tab:branch-scan}
\begin{tabular}{crr}
\toprule
$B$ & Test NRMSE & OOD $n/m/h$ $R^2$ \\
\midrule
1 & 0.529 & 0.808/0.555/0.663 \\
2 & 0.469 & 0.843/0.670/0.808 \\
3 & 0.385 & 0.916/0.696/0.811 \\
4 & 0.445 & 0.711/0.678/0.457 \\
5 & 0.424 & 0.779/0.778/0.650 \\
\bottomrule
\end{tabular}
\end{table}

Transient weighting $w$ reallocates learning toward observable fast events,
but its effect depends on topology. Across five seeds, prediction has a broad
$w=1$--$2$ plateau. The constrained two-branch model has its strongest
physical state metrics at $w=2$, while five branches are best regularized at
$w=1$. None removes the fast-$m$ transported-field gap under the
fitted-chart audit.

\begin{table*}[t]
\centering
\caption{Five-seed topology--loss operating points. Gate columns are strict
OOD $R^2$; the scale column is the fitted descriptive
$\tau_m^{\mathrm{eff}}/\tau_m^{\mathrm{HH}}$.}
\label{tab:topology-loss}
\begin{tabular}{ccrrrrrr}
\toprule
$B$ & $w$ & NRMSE & $n$ & $m$ & $h$ & $\tau_m^{\mathrm{eff}}$ scale & $m$-field $R^2$ \\
\midrule
2 & 1 & 0.3411 & 0.927 & 0.901 & 0.899 & $6.00\times$ & 0.175 \\
2 & 2 & 0.3416 & 0.929 & 0.904 & 0.918 & $5.18\times$ & 0.238 \\
5 & 1 & 0.3287 & 0.933 & 0.863 & 0.866 & $5.96\times$ & 0.091 \\
5 & 2 & 0.3307 & 0.909 & 0.847 & 0.846 & $5.84\times$ & 0.103 \\
\bottomrule
\end{tabular}
\end{table*}

Reliability separates otherwise similar operating points. Across five
independently trained checkpoints per setting, with two chart fits per
checkpoint, $B=5,w=1$ has far-event $m$-field RMSE
$0.237\pm0.044$ with 0/10 runs above 0.5, whereas $B=5,w=2$ gives
$0.384\pm0.343$ with 2/10 runs above 0.5. The $B=2,w=1$ and $B=2,w=2$
groups give $0.997\pm0.762$ and $0.947\pm0.733$, each with 6/10 runs above
0.5, despite far-event state $R^2\simeq0.9996$--$0.9998$. Selection using
held-out ID far-event total-field or $m$-specific error does not repair this
tail. Its decomposition combines generator-amplified state error,
chart-derivative residual, and variable cancellation.

\begin{figure}[H]
  \centering
  \includegraphics[width=\textwidth]{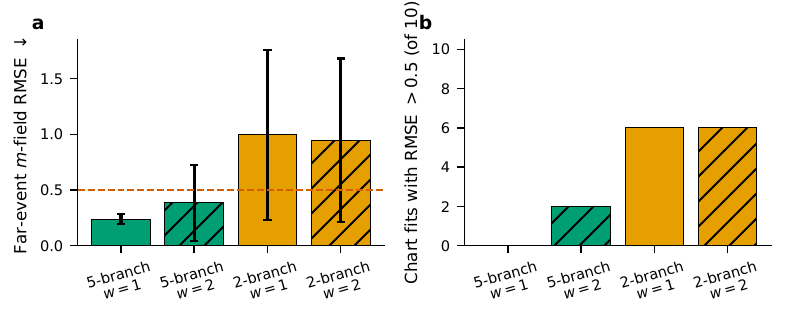}
  \caption{Branch redundancy improves chart-fit reliability while the fast-$m$
  transported-field gap remains. (a) Far-event $m$-field relative RMSE per
  topology--loss operating point (bars, mean $\pm$ sample SD over ten chart
  fits, five training seeds times two chart initializations; dashed line,
  $0.5$). (b) Number of those ten fits above $0.5$.
  Table~\ref{tab:topology-loss} lists the corresponding operating points.}
  \label{fig:field-reliability-appendix}
\end{figure}

\subsection{Robustness and additional-coordinate audits}
\label{app:robustness}

Forty-millisecond rollout remains stable, but total NRMSE rises by 3--16\%,
current-clamp NRMSE by 15--23\%, and spike-exact rate falls by 15--36
percentage points. Ten-percent HH parameter perturbations increase aggregate
error by 5--13\%, with larger peak-current errors. These tests support
bounded protocol and horizon transfer, not a parameter-invariant HH law.

With 10\% response noise and tenfold history sparsity, $K=3/K=4$ NRMSE is
$0.457/0.473$; at 20\% and 25-fold it is $0.480/0.501$. On matched windows,
40\% noise and 50-fold sparsity raises NRMSE from $0.443$ to $0.524$ for
$K=3$ and from $0.451$ to $0.546$ for $K=4$. A chart fitted only on clean ID
data remains transferable: for the three transient-weighted $K=3$
checkpoints (seeds 56--58), OOD $n/m/h$ scores decline from
$0.872/0.735/0.767$ to $0.832/0.700/0.739$ at the extreme condition. These
checkpoints come from the transient-loss study rather than the locked
dimension scan of Figure~\ref{fig:dimension-transitions} and use a different
selected ridge, so we compare within-study trends rather than absolute
cross-table values.

Nonlinear grouped reconstruction explains the overcomplete coordinates from
finite observable history. In voltage clamp, physical state plus history
reconstructs $K=3/K=4$ latent state with variance-weighted
$R^2=0.984/0.991$. Using only $I/V$ history still obtains $0.975/0.986$;
the $K=4$ result is 0.982/0.987/0.990 over three checkpoints. Current-clamp
observable-history scores are 0.850/0.812. Together with PCA, this is consistent
with a history-derived disentangling role within the sampled protocol support.

\subsection{Additional design controls}
\label{app:negative-results}

We examined direct long rollout,
multihorizon curricula, active-protocol augmentation, larger encoders, global
and conditional spike losses, task reweighting, latent group penalties, and
linear through cubic charts. Direct 20-ms training, active voltage protocols,
adequate history, and moderate capacity improved prediction. Across these
design controls, the fast-$m$ transported-field gap persists.

Dynamics-aware chart optimization shows a reproducible empirical state--field
trade-off:
more field weight reduces transported-field error while lowering state
$R^2$. Across independent starts and stabilized training, the empirical
state--field trade-off persists. On the same support, chart complexity moves state and field agreement
in the same direction: a linear chart lowers $m$-state $R^2$ from $0.992$ to
$0.941$ and $m$-field $R^2$ from $0.271$ to $0.039$ at $K=3$
(Table~\ref{tab:chart-classes}).
Table~\ref{tab:exclusion-chain} collects these design choices and what each
control showed.

\begin{table}[H]
\centering
\caption{Design choices examined for the fast-$m$ comparison. Each row varies one aspect of predictor training or post-lock chart fitting and reports the resulting change.}
\label{tab:exclusion-chain}
\begin{tabular}{>{\RaggedRight\arraybackslash}p{0.20\textwidth}>{\RaggedRight\arraybackslash}p{0.32\textwidth}>{\RaggedRight\arraybackslash}p{0.38\textwidth}}
\toprule
Design choice & Control & Observation \\
\midrule
Optimization budget & Independent starts, longer and stabilized dynamics-aware chart training & The empirical state--field trade-off persists across the tested settings. \\
Chart capacity & Four-layer, 256-unit, 20,000-step chart & Repeated resampling leaves the empirical state--field trade-off intact. \\
Chart complexity & Linear through cubic charts and grouped-CV regularization & On the same support a linear chart lowers both state and field agreement (Table~\ref{tab:chart-classes}), so the gap is not unique to the cubic chart. \\
Voltage chain rule & Voltage-conditioned and clamp-specific charts with the full $\partial_Vr\,\dot V$ term & Accounts for part of the early discrepancy but not the voltage-clamp fast-$m$ field error. \\
Transient emphasis & Jump-stratified sampling and event-loss weights & Moves the selected state--field operating point without removing the fast-$m$ transported-field gap. \\
History, horizon, and excitation & Longer histories, direct multihorizon rollout, and active voltage protocols & Improves prediction and audit support; the transported-field gap remains under the tested audit. \\
\bottomrule
\end{tabular}
\end{table}

\paragraph{Observable training-design controls.}
Directly training the requested 20-ms horizon improves on 5--10-ms training
followed by extrapolation. Increasing history from 10 to 15 ms improves mean
NRMSE from $0.4664\pm0.0060$ to $0.4514\pm0.0026$, spike-count exact rate
from $28.84\%\pm0.52\%$ to $31.76\%\pm0.36\%$, and first-spike timing
MAE from $0.765\pm0.088$ to $0.584\pm0.067$ ms over five paired seeds.
Twenty-millisecond history improves some current-clamp events but degrades
voltage-clamp prediction. Spike-focused resampling and global or conditional
spike losses improve selected peak metrics only by sacrificing waveform error.

\subsection{Claim-to-evidence map}
\label{app:evidence-map}

Table~\ref{tab:evidence-map} maps each claim to the experiment family that
supports it, the independent units behind it, and where it is reported.

\begin{table}[H]
\centering
\caption{Summary of the experimental evidence supporting the main findings.}
\label{tab:evidence-map}
\small
\begin{tabular}{>{\RaggedRight\arraybackslash}p{0.25\textwidth}>{\RaggedRight\arraybackslash}p{0.19\textwidth}>{\RaggedRight\arraybackslash}p{0.18\textwidth}>{\RaggedRight\arraybackslash}p{0.18\textwidth}}
\toprule
Claim & Experiment families & Independent units & Paper location \\
\midrule
Prediction error and seed spread drop sharply at $K=3$ & Latent-dimension scan, $K=1$--$6$ & Five training seeds, $K=1$--$6$ & Figure~\ref{fig:dimension-transitions}; Table~\ref{tab:dimension-complete} \\
Three-dimensional covariance core in 13 of 15 overcomplete models & Latent covariance and geometry audits & $K=4$--$6$, five seeds per $K$ & Figure~\ref{fig:dimension-transitions}; Appendix~\ref{app:dimension} \\
Protocol alignment improves through $K=5$--$6$ & Locked in-distribution-to-OOD chart transfer & Five seeds per $K=3$--$6$, grouped-CV repeats & Figure~\ref{fig:dimension-transitions}; Table~\ref{tab:dimension-complete} \\
State recovery does not certify the generator & Gate-state and transported-field audits & Gate-wise, cross-seed, analytic and numerical controls & Figures~\ref{fig:state-generator-gap} and \ref{fig:mechanism-diagnostics} \\
Redundancy changes reliability, not correctness & Branch-count and transient-loss factorial & Five training seeds and two chart starts & Table~\ref{tab:topology-loss}; Figure~\ref{fig:field-reliability-appendix} \\
Gate information transfers across tested architectures & Baseline architecture comparisons & Three seeds per baseline family & Table~\ref{tab:baseline-main}; Figure~\ref{fig:baseline-context} \\
Numerical and observational robustness & Simulator validation; long-horizon, parameter, and noisy or sparse observation controls & Independent solver, horizons, parameters, noise, and sparsity & Appendices~\ref{app:sim-validation} and \ref{app:robustness} \\
\bottomrule
\end{tabular}
\end{table}